\pdfoutput=1

\documentclass[11pt]{article}

\usepackage[final]{acl}

\usepackage{times}
\usepackage{latexsym}

\usepackage[T1]{fontenc}

\usepackage[utf8]{inputenc}

\usepackage{microtype}

\usepackage{inconsolata}

\usepackage{graphicx}
\usepackage{subcaption}
\usepackage{eso-pic} 

\usepackage{amsmath}
\usepackage{amsfonts}

\usepackage{multirow}
\usepackage{booktabs}
\usepackage{siunitx}
\usepackage{makecell}

\usepackage{url}

\usepackage{adjustbox} 

\usepackage{pifont} 

\usepackage{arydshln} 

\usepackage{booktabs, siunitx, multirow, xcolor}

\usepackage{threeparttable}

\definecolor{bestblue}{RGB}{0, 102, 204}

\title{Probabilistic Textual Time Series Depression Detection}

\author{
 \textbf{Fabian Schmidt\textsuperscript{1}},
 \textbf{Seyedehmoniba Ravan\textsuperscript{2}},
\textbf{Vladimir Vlassov\textsuperscript{1}}
\\
\\
 \textsuperscript{1}Department of Computer Science, KTH Royal Institute of Technology, Sweden\\
 \textsuperscript{2}Department of Information Technology, Uppsala University, Sweden\\
 \small{
   \textbf{Correspondence:} \href{mailto:fschm@kth.se}{fschm@kth.se}
 }
}

\begin{document}
\AddToShipoutPictureFG*{%
  \put(0,0){%
    \parbox[b][\paperheight][b]{\paperwidth}{%
      \centering
      {\footnotesize
        \textit{Findings of the Association for Computational Linguistics: ACL 2026}, pages 32574--32589\\[1pt]
        July 2-7, 2026 \textcopyright{}2026 Association for Computational Linguistics\par}%
      \vspace{28pt}%
    }%
  }%
}%
\maketitle

\begin{abstract}
Accurate and interpretable predictions of depression severity are essential for clinical decision support, yet existing models often lack uncertainty estimates and temporal interpretability. We propose PTTSD, a \emph{Probabilistic} framework for \emph{Depression Detection} from clinical interview utterance sequences that predicts PHQ-8 scores while modeling calibrated uncertainty. PTTSD includes sequence-to-sequence and sequence-to-one variants, both combining LSTMs, self-attention, and residual connections with Gaussian or Student's-$t$ output heads trained via negative log-likelihood. The sequence-to-sequence variant enables temporal analysis of how predictive confidence evolves over an interview, despite the target being a single session-level score. Evaluated on E-DAIC and DAIC-WOZ, PTTSD achieves competitive performance among text-only systems (e.g., MAE = 3.85 on E-DAIC, 3.55 on DAIC) and produces well-calibrated prediction intervals. Ablations confirm the value of attention and probabilistic modeling, while a three-part calibration analysis and qualitative case studies highlight the clinical relevance of uncertainty-aware prediction.
\end{abstract}

\section{Introduction}

Depression remains one of the leading causes of global disability, affecting over 300 million individuals worldwide~\cite{who2017depression,who2022mentalhealth}. Scalable, automated tools for assessing depressive symptom severity offer valuable support in digital therapy and remote care, where access to clinicians is limited. Among these tools, text-based systems that process clinical interviews have shown strong potential for predicting standardized scores such as the PHQ-8 \cite{kroenke2009phq}.

Recent methods typically model interview transcripts as sequences of utterances and employ architectures such as LSTMs, Transformers, or large language models (LLMs)~\cite{mandal-etal-2025-enhancing, fang2023multimodal, nykoniuk2025multimodal, sadeghi2024harnessing}. However, most existing approaches produce scalar severity estimates without quantifying uncertainty, which is an important limitation in high-stakes clinical contexts where a prediction of ``PHQ-8 = 12'' is far more actionable when accompanied by a measure of confidence. 

We argue that the sequential nature of clinical interviews creates a natural opportunity to address this gap. Each utterance provides a context-dependent observation, and the cumulative sequence progressively constrains the space of plausible severity estimates. While the prediction target, i.e., the PHQ-8 score, is a single-session-level value rather than a time-varying quantity, the \emph{input} is inherently sequential, and modeling it as such offers two key advantages that point-estimate systems forgo. First, it allows probabilistic models to capture \emph{aleatoric uncertainty}, that is, the uncertainty arising from sparse, contradictory, or ambiguous language, and to express how that uncertainty resolves as context accumulates. Second, it enables interpretable temporal analyses: identifying which utterances drive prediction shifts, and how model confidence stabilizes (or fails to stabilize) over the course of an interview.

We introduce \textit{PTTSD}, a \emph{Probabilistic Textual Time Series Depression Detection} model that makes temporally grounded, calibrated predictions over PHQ-8 scores from utterance-level sequences. PTTSD addresses two key gaps in the field. First, it replaces point predictions with calibrated distributional outputs (Gaussian or Student's-$t$ heads trained via negative log-likelihood), enabling clinicians to assess both predicted severity and the model's confidence in that prediction. Second, through its sequence-to-sequence (seq-to-seq) variant, it exposes how the model's predictive belief evolves across an interview, providing temporal interpretability that is absent from prior systems. Clinicians can therefore identify when the model becomes confident, which utterances drive prediction shifts, and where ambiguity persists. 

We evaluate PTTSD on the DAIC and E-DAIC benchmarks using original and re-transcribed interviews and demonstrate competitive performance on standard metrics (e.g., MAE = 3.55, RMSE = 4.77 on DAIC; MAE = 3.85, RMSE = 4.52 on E-DAIC), matching or exceeding recent text-only baselines on held-out test sets without relying on prompt engineering or handcrafted features. Importantly, PTTSD additionally provides calibrated uncertainty estimates and interpretable temporal dynamics—capabilities absent from prior systems. Ablation and sensitivity analyses further validate the contributions of probabilistic loss design, attention mechanisms, and calibration metrics.

In summary, our main contributions are:
\begin{itemize}
    \item We propose PTTSD, a fully probabilistic sequence model that jointly predicts PHQ-8 scores along with calibrated uncertainty from utterance-level textual time series, and provide thorough calibration, temporal, and sensitivity analyses to assess uncertainty quality and clinical relevance.
   \item We introduce a seq-to-one and a seq-to-seq formulation. The latter exposes how the model's predictive belief, both point estimate and uncertainty, evolves over the course of an interview for temporal interpretability analyses such as identifying critical utterances and tracking when model confidence stabilizes.
    \item We empirically demonstrate competitive results on E-DAIC and DAIC test sets among text-only models, while offering calibrated interpretable uncertainty estimates that go beyond the point predictions of prior work.
\end{itemize}

\section{Related Work}\label{sec:related_work}

Textual time series modeling has been central to recent efforts in automatic depression detection, especially within clinical interviews and therapy sessions. Prior work has predominantly relied on point estimate neural methods such as LSTMs and attention-based transformers to model temporal dependencies in textual data~\cite{mandal-etal-2025-enhancing, fang2023multimodal, nykoniuk2025multimodal}. Such models capture sequential patterns but lack mechanisms to quantify temporal uncertainty. While LLMs extract richer textual features~\cite{sadeghi2024harnessing, chen-etal-2024-depression}, most systems remain heuristic or point estimates that focus on structural or multimodal fusion rather than probabilistic reasoning. In contrast, our fully probabilistic, end-to-end model captures uncertainty directly from raw utterances without handcrafted prompts.

Notably, \citet{qureshi2019multitask} use multitask learning with attention mechanisms for joint regression and classification, but do not incorporate uncertainty modeling. Similarly, prompt-based methods such as those of \citet{zhang2024prompt} transform depression detection into a few-shot classification task via language model prompting, but still yield single-point predictions. Graph-based architectures~\cite{burdisso2023graph, chen-etal-2024-depression} model discourse-level context across utterances and questions, offering enhanced interpretability and structural awareness, though they too typically omit calibrated uncertainty.

A rare exception is \citet{dia2024paying}, who propose a stochastic transformer for post-traumatic stress disorder detection, introducing probabilistic components such as stochastic activations to model uncertainty across modalities. However, their work focuses on visual signals and does not address textual time series or PHQ-8 regression. More recently, \citet{zhang2025mil} apply a multi-instance learning (MIL) framework to estimate depression severity from long transcripts and assign confidence scores to depressive cues at the sentence level. This approach does provide instance-level interpretability, but the underlying model is not explicitly probabilistic in the Bayesian sense.

Several recent works have explored fair or calibrated uncertainty estimation. \citet{li2025fair} propose Fair Uncertainty Quantification (FUQ) for PHQ regression with conformal prediction intervals across demographic groups. While effective for fairness, FUQ operates at the distributional output level and does not model temporal evolution within interviews. Other systems, such as \citet{mao2022prediction} and \citet{guo2022topic}, employ BiLSTMs or Transformers with textual features, sometimes augmented by topic signals, but focus solely on point estimate loss objectives.

\section{Probabilistic Textual Time-Series Depression Detection}
\label{sec:methods}

\begin{figure*}[!t]
    \centering
    \includegraphics[width=.7\textwidth]{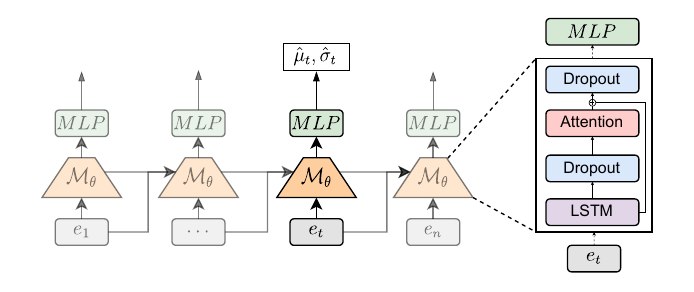}
    \caption{Overview of PTTSD. Utterances are embedded, encoded with a BiLSTM and multi-head self-attention with residual connections, and mapped to a predictive distribution over PHQ-8.}
    \label{fig:the_method}
\end{figure*}

\subsection{Problem Formulation}
\label{sec:problem_formulation}
We model PHQ-8 estimation as probabilistic regression over utterance sequences. Given a transcript with $T$ utterances $\{u_1,\dots,u_T\}$ and utterance embeddings $e_{1:T}$, the model maps the sequence to a distribution over the session-level score $y\in\mathbb{R}$:
\[
p\!\left(y \mid e_{1:T}; \theta\right).
\]

We study two PTTSD model variants:
\begin{itemize}
    \item \emph{seq-to-one}: predicts a single distribution from the full utterance sequence.
    \item \emph{seq-to-seq}: produces a per-utterance distribution $p(y \mid e_{1:t}; \theta)$ for each prefix $e_{1:t}$, trained against the same session-level label $y$.
\end{itemize}

We note that the PHQ-8 target is a single score per session, not a time-varying quantity. The seq-to-seq variant therefore does not model a changing target. Rather, it exposes how the model's \emph{predictive belief}, i.e., its point estimate $\hat{\mu}_t$ and its uncertainty $\hat{\sigma}_t$, evolves as conversational context accumulates. This serves two purposes: (i) it enables temporal interpretability analyses, such as identifying utterances that trigger large shifts in predicted severity or uncertainty, and (ii) it acts as a form of regularization, encouraging the model to form reasonable estimates from partial context rather than relying solely on global sequence features.

Concretely, the trajectories of $(\hat{\mu}_t, \hat{\sigma}_t)$ across an interview support three types of analysis: (i) how uncertainty decreases as context accumulates and signals that the model has ``seen enough''; (ii) error and uncertainty alignment to measure whether high predicted uncertainty genuinely corresponds to high error (in terms of correlations and interval coverage); and (iii) identification of critical utterances where the predicted mean shifts sharply or attention mass concentrates, which may correspond to clinically salient moments in the interview.

\subsection{Data}
\label{sec:data}
We utilize the Distress Analysis Interview Corpus (DAIC) \cite{gratch-etal-2014-distress} and extended DAIC (E-DAIC) \cite{edaic2020} datasets, which contain anonymized semi-structured interview transcripts and associated Patient Health Questionnaire-8 (PHQ-8) \cite{kroenke2009phq} depression scores. Each participant's data consists of a sequence of utterances extracted from transcript files, along with a PHQ-8 score indicating depression severity. 
The PHQ-8 is a standardized self-report instrument with scores ranging from 0 to 24 that assesses depressive symptom severity. The DAIC‑WOZ corpus includes 189 clinical interview sessionss. Its extended counterpart, E‑DAIC, includes 275 sessions. Both remain among the few publicly available conversation‑based clinical corpora annotated with PHQ‑8 depression scores, collected under stringent ethical and privacy safeguards typical of mental health research. More details on the PHQ-8 and DAIC in Appendix~\ref{app:phq8} and Appendix~\ref{app:edaic}, respectively.

\subsection{Utterance Embeddings}
We represent each utterance using pretrained sentence encoders. Our primary model uses the \texttt{all-MiniLM-L6-v2} Sentence Transformer\footnote{\url{https://huggingface.co/sentence-transformers/all-MiniLM-L6-v2}}~\cite{reimers-gurevych-2019-sentence}, a compact model with only 22 million parameters that achieves competitive performance across a wide range of tasks on the Hugging Face MTEB Embedding Leaderboard~\cite{muennighoff-etal-2023-mteb}. We also evaluate an alternative variant of our model using \texttt{MentalBERT}~\cite{ji-etal-2022-mentalbert}, a domain-adapted BERT model pretrained on mental health-related corpora\footnote{\url{https://huggingface.co/mental/mental-bert-base-uncased}}. Tokenization, pooling, and other encoder specifics are in Appx.~\ref{app:impl}.

\subsection{Backbone: BiLSTM + Self-Attention}
\label{sec:backbone}
We employ a multi-layer unidirectional LSTM that encodes $\mathbf{X}$ into hidden states $\mathbf{H}\in\mathbb{R}^{T\times H}$. We then apply multi-head self-attention with a residual connection:
\[
\mathbf{A} \;=\; \textsc{Attention}(\mathbf{H}) \;+\; \mathbf{H},
\]
allowing each utterance embedding to condition on the full conversational context. In \emph{seq-to-one}, we aggregate $\mathbf{A}$ over time by average pooling. We retain the per-utterance representations $\mathbf{a}_t$ in \emph{seq-to-seq}.

\subsection{Uncertainty-Aware Output Heads}
Separate MLP heads predict the parameters of the predictive distribution to model the PHQ-8 score ($\hat{\mu}$) and uncertainty ($\hat{\sigma}$).

\[
p(y\mid\cdot)\in\left\{\mathcal{N}\!\big(\hat{\mu},\hat{\sigma}^2\big),\; \text{Student-}t\!\big(\hat{\mu},\hat{\sigma},\nu\big)\right\}.
\]
For \emph{seq-to-one}, the heads take the pooled vector as input and for \emph{seq-to-seq}, the heads take each $\mathbf{a}_t$ to predict the parameters at each timestep. 

\subsection{Training Objective}
\label{sec:losses_main}
We minimize the NLL of the ground-truth PHQ-8 under the predicted distribution. For \emph{seq-to-one}:
\[
\mathcal{L} \;=\; -\log p\!\left(y \mid e_{1:T};\theta\right).
\]
For \emph{seq-to-seq}, we average the per-utterance NLL across the sequence. Gaussian vs.\ Student’s-$t$ objectives and optional weighting terms are provided in Appx.~\ref{app:nll}. Optimization settings and other training hyperparameters are in Appx.~\ref{app:impl}.

\section{Experiments}\label{sec:experiments}

\subsection{Experimental Setup}

\paragraph{Data Splits.}
We follow the official training, validation, and test splits provided with each dataset. For E-DAIC, the data is partitioned into 163 training, 56 validation, and 56 test participants. For DAIC-WOZ, the official splits include 107 training, 35 validation, and 56 test participants. E-DAIC audio is re-transcribed using WhisperX to improve transcription quality and alignment over the original transcripts to allow for better uncertainty quantification, which is the main focus of our study.
To ensure fair comparison with prior work that used the original transcripts, we additionally report results on the unaltered E-DAIC transcripts in Appendix~\ref{app:original_transcripts}.

\paragraph{Evaluation Metrics.}
We evaluate models on both the validation and held-out test sets using mean squared error (MSE) and root mean squared error (RMSE). MSE and RMSE quantify average prediction error, with RMSE placing greater emphasis on larger errors due to its squaring operation. RMSE is particularly useful for identifying models that minimize not only average error but also the variance in error magnitude. When modeling predictive uncertainty, we additionally report negative log-likelihood (NLL).

\paragraph{Reproducibility.}
All preprocessing steps, model configurations, and training scripts are made publicly available on GitHub.\footnote{\url{https://github.com/smidtfab/PTTSD}
} To account for variability due to random initialization, we report average performance over three runs with different seeds.

\subsection{Main Results}

\begin{table*}[!ht]
\centering
\scriptsize
\setlength{\tabcolsep}{4pt}

\begin{subtable}[t]{0.49\textwidth}
\vspace{0pt}
\centering
\resizebox{\linewidth}{!}{%
\begin{tabular}{lcccc}
\toprule
& \multicolumn{2}{c}{\textbf{Dev}} & \multicolumn{2}{c}{\textbf{Test}} \\
\cmidrule(lr){2-3} \cmidrule(lr){4-5}
\textbf{Method} & \textbf{MAE} & \textbf{RMSE} & \textbf{MAE} & \textbf{RMSE} \\
\midrule
\citet{williamson2016speech}   & 3.34          & 4.46          & --            & --            \\
\citet{gong2020topic}          & \textbf{2.77} & \textbf{3.54} & 3.96          & 4.99          \\
\citet{yang2016context}        & 3.52          & 4.52          & --            & --            \\
\citet{stepanov2021multimodal} & --            & --            & 4.88          & 5.83          \\
\citet{oureshi2017deep}        & 3.78          & --            & --            & --            \\
\citet{niu2019hierarchical}    & 3.73          & 4.80          & --            & --            \\
\citet{fang2023transformer}    & --            & --            & 3.61          & 4.76          \\
\citet{rohanian2019contextual} & --            & --            & 4.98          & 6.05          \\
\citet{alhanai2018sequence}    & 5.18          & 6.38          & --            & --            \\
\citet{qureshi2017ensemble}    & 3.74          & 4.80          & --            & --            \\
\midrule
PTTSD seq-to-one (MentalBERT) & 4.39{\tiny$\pm$0.10} & 5.47{\tiny$\pm$0.43} & 3.65{\tiny$\pm$0.24} & \textbf{4.69}{\tiny$\pm$0.24} \\
PTTSD seq-to-seq (MentalBERT) & 4.67{\tiny$\pm$0.04} & 5.82{\tiny$\pm$0.34} & 3.92{\tiny$\pm$0.54} & 4.79{\tiny$\pm$0.54} \\
PTTSD seq-to-one (MiniLM)     & 3.82{\tiny$\pm$0.09} & 4.84{\tiny$\pm$0.28} & \textbf{3.55}{\tiny$\pm$0.15} & 4.77{\tiny$\pm$0.53} \\
PTTSD seq-to-seq (MiniLM)     & 4.59{\tiny$\pm$0.07} & 5.22{\tiny$\pm$0.30} & 3.88{\tiny$\pm$0.41} & 5.10{\tiny$\pm$0.92} \\
\bottomrule
\end{tabular}}
\caption{DAIC-WOZ}
\label{tab:DAIC_text_results}
\end{subtable}
\hfill
\begin{subtable}[t]{0.49\textwidth}
\vspace{0pt}
\centering
\resizebox{\linewidth}{!}{%
\begin{tabular}{lcccc}
\toprule
& \multicolumn{2}{c}{\textbf{Dev}} & \multicolumn{2}{c}{\textbf{Test}} \\
\cmidrule(lr){2-3} \cmidrule(lr){4-5}
\textbf{Method} & \textbf{MAE} & \textbf{RMSE} & \textbf{MAE} & \textbf{RMSE} \\
\midrule
\citet{ray2019multilevel}                           & --            & 4.37          & 4.02          & 4.73          \\
\citet{makiuchi2019multimodal} LSTM                 & --            & 4.97          & --            & 6.88          \\
\citet{makiuchi2019multimodal} LSTM+CNN & --            & 4.22          & --            & --            \\
\citet{sadeghi2023depression}                       & 3.65          & 5.27          & 4.26          & 5.37          \\
\citet{sadeghi2024harnessing} Pr3+Whisper           & 3.17          & 4.51          & 4.22          & 5.07          \\
\citet{sadeghi2024harnessing} Pr3+Whisper+AQ & \textbf{2.85} & \textbf{4.02} & 3.86          & 4.66          \\
\midrule
PTTSD seq-to-one (MentalBERT)$^{\dagger}$ & 3.56{\tiny$\pm$0.01} & 4.45{\tiny$\pm$0.07} & 4.18{\tiny$\pm$0.05} & 5.23{\tiny$\pm$0.13} \\
PTTSD seq-to-seq (MentalBERT)$^{\dagger}$ & 3.55{\tiny$\pm$0.14} & 4.58{\tiny$\pm$0.20} & 4.20{\tiny$\pm$0.03} & 5.39{\tiny$\pm$0.08} \\
PTTSD seq-to-one (MiniLM)$^{\dagger}$     & 3.60{\tiny$\pm$0.13} & 4.76{\tiny$\pm$0.14} & 4.58{\tiny$\pm$0.50} & 5.87{\tiny$\pm$0.92} \\
PTTSD seq-to-seq (MiniLM)$^{\dagger}$     & 3.47{\tiny$\pm$0.02} & 4.57{\tiny$\pm$0.04} & \textbf{3.85}{\tiny$\pm$0.04} & \textbf{4.52}{\tiny$\pm$0.38} \\
\bottomrule
\end{tabular}}
{\scriptsize\raggedright $^{\dagger}$\,PTTSD results obtained on WhisperX re-transcriptions of the original E-DAIC audio, similar to \citet{sadeghi2024harnessing}'s approach rather than transcripts from the dataset. See Appendix~\ref{app:original_transcripts} for results on the original transcripts.\par}
\par\vspace{.93em}
\caption{E-DAIC}
\label{tab:EDAIC_text_results}
\end{subtable}

\caption{Comparison of text-only PHQ-8 regression models, split by dataset. Bold = best.}
\label{tab:E-and-DAIC_text_results}
\end{table*}

Table~\ref{tab:E-and-DAIC_text_results} presents PHQ-8 regression performance on both E-DAIC and DAIC. We compare our PTTSD models across multiple configurations (seq-to-seq vs. seq-to-one and MentalBERT vs. all-MiniLM-L6-v2) against relevant text-based approaches.

\paragraph{E-DAIC.} Among the text-only systems evaluated, PTTSD (seq-to-seq with all-MiniLM-L6-v2) achieves the lowest test MAE (3.85) and RMSE (4.52) in our comparison. Other PTTSD variants, including MentalBERT-based and seq-to-one configurations, also perform competitively, showing robustness across architecture choices.
We note that direct numerical comparison with prior work is complicated by two factors: (i) our use of WhisperX re-transcriptions may yield different input quality than the original transcripts used by prior systems, and (ii) several baselines report only development or only test metrics, making comprehensive comparison difficult.
Earlier works such as \citet{ray2019multilevel} and \citet{makiuchi2019multimodal} attain dev RMSEs of 4.22–4.97, but their test performance is either weaker or unreported. More recent prompt-based models by \citet{sadeghi2024harnessing} use Whisper transcripts and audio-based quality filtering. Their best variant (Pr3+Whisper+AudioQual) reports strong dev MAE (2.85) and RMSE (4.02) with the additional audio quality gating. Their text-only variant (Pr3+Whisper) achieves 4.22 MAE and 5.07 RMSE on the test set, which PTTSD improves upon on both metrics. However, given the differences in the transcripts, these comparisons should be interpreted with caution.
To directly address comparability, we also evaluate PTTSD on the original (unaltered) E-DAIC transcripts. Using seq-to-seq with Gaussian NLL, MiniLM achieves a test MAE of 4.63 and RMSE of 5.56, while MentalBERT achieves 4.60 MAE and 5.58 RMSE (Appendix~\ref{app:original_transcripts}, Table~\ref{tab:original_transcripts}). While these numbers are expectedly weaker than those obtained on re-transcribed data, they confirm that PTTSD remains competitive and that the core contributions, the calibrated uncertainty and interpretable temporal trajectories, hold regardless of transcript source and are arguably more valuable for clinical deployment than marginal improvements in MAE or RMSE.

\paragraph{DAIC.} On the original DAIC dataset, PTTSD again performs competitively, especially in the all-MiniLM-L6-v2 seq-to-one variant, which achieves the lowest test MAE (3.55) and matches the best test RMSE (4.77) of \citet{fang2023multimodal}. Interestingly, \citet{gong2020topic} reports strong dev performance (MAE 2.77, RMSE 3.54), while test results (MAE 3.96, RMSE 4.99) show a notable drop, which may reflect differences in evaluation protocols or generalization challenges.

\paragraph{Seq-to-seq vs.\ seq-to-one.}
The seq-to-seq variant (MiniLM) outperforms seq-to-one on E-DAIC, whereas the pattern reverses on DAIC. We do not interpret this as evidence that one formulation is strictly superior. Rather, the two variants serve complementary roles. Seq-to-one optimizes for aggregate prediction quality from the full sequence, while seq-to-seq provides temporal interpretability at a modest and dataset-dependent cost to accuracy. The seq-to-one variant uses masked average pooling, whereas seq-to-seq retains per-timestep representations. Exploring alternative pooling strategies (e.g., attention-weighted aggregation) is a viable direction for future work.

\subsection{Ablation Studies} 

\paragraph{Effect of Loss Function.}

\begin{table}[ht]
\centering
\footnotesize
\begin{tabular}{lcc|cc}
\toprule
\textbf{Loss} & \multicolumn{2}{c|}{\textbf{Dev}} & \multicolumn{2}{c}{\textbf{Test}} \\
              & \textbf{MAE} & \textbf{RMSE}      & \textbf{MAE} & \textbf{RMSE} \\
\midrule
Gaussian NLL     & 3.4440 & 4.5293 & 3.8603 & 5.0219 \\
Student-$t$ NLL  & 3.6637 & 4.9328 & 3.9294 & 5.1488 \\
MAE              & 3.6427 & 4.8091 & 4.1885 & 5.4407 \\
MSE              & 3.6398 & 4.9845 & 3.6694 & 4.8760 \\
\bottomrule
\end{tabular}
\caption{Loss function comparison on dev/test sets (E-DAIC, single run).}
\label{tab:loss_comparison}
\end{table}

Table~\ref{tab:loss_comparison} compares the impact of different loss functions on validation and test performance. Gaussian NLL yields the best overall balance with low MAE and RMSE across both splits, particularly on test MAE (3.86). Student's-$t$ NLL performs comparably but with slightly worse calibration and higher RMSE, likely due to the added complexity of estimating the degrees of freedom.
MAE and MSE losses exhibit inconsistent behavior. While MSE achieves the lowest test MAE (3.67), it performs worse on the dev set and yields the highest test RMSE among all probabilistic losses. The MAE loss underperforms across all metrics, suggesting it is less effective at learning stable sequence-level representations in this setting.
In summary, Gaussian NLL offers the most reliable and generalizable performance when modeling uncertainty in PHQ-8 prediction from textual time series.

\paragraph{Effect of the Model Architecture.}
We conduct an ablation study to assess the contribution of individual architectural components in our probabilistic LSTM seq-to-seq model. Each ablation variant disables a specific component (attention, residual connections, or the variance prediction head) while all other settings are held constant. The models are trained for 50 epochs with early stopping (patience of 15 epochs). The final evaluation is performed on the test set using MAE, RMSE, and NLL, averaged over three random seeds. Full experimental details are included in Appendix~\ref{app:ablation}.

\begin{table}[ht]
\centering
\resizebox{\columnwidth}{!}{%
\begin{tabular}{lccccccc}
\toprule
\textbf{Variant} & \textbf{MAE} & \textbf{$\Delta$ MAE (\%)} & \textbf{RMSE} & \textbf{$\Delta$ RMSE (\%)} & \textbf{NLL} \\
\midrule
Full Model           & 3.85 & --      & 4.99 & --      & 1.05 \\
- w/o Attention      & 5.91 & +53.29  & 7.40 & +48.28  & 1.56 \\
- w/o Residual       & 5.34 & +38.52  & 6.66 & +33.42  & 1.43 \\
- w/o Variance Head  & 3.99 & +3.57   & 5.24 & +5.04   & --   \\
\bottomrule
\end{tabular}%
}
\caption{Ablation of architectural components averaged over three seeds. Absolute scores and percentage change relative to the full model. NLL is not applicable (--) for the variant without a variance head, as it produces only point estimates.}
\label{tab:ablation_attention}
\end{table}

Table~\ref{tab:ablation_attention} illustrates the effects of disabling each component. Self-attention produces the largest degradation when removed, with MAE increasing by 53.3\% and RMSE by 48.3\%. This is expected given that interviews span up to 355 utterances (median: 164). Such a range over which recurrent hidden states alone can be insufficient to maintain coherent long-range context. Omitting residual connections also causes substantial performance drops (MAE +38.5\%, RMSE +33.4\%). Hence, skip connections are essential for stable gradient flow and effective feature reuse across layers. Ablating the variance prediction head degrades performance across all metrics, including raw point-estimate accuracy (MAE +3.6\%, RMSE +5.0\%), and results in the expected loss of calibrated uncertainty estimates. We conjecture that probabilistic training acts as a form of regularisation and that by explicitly modeling aleatoric uncertainty, the model avoids over-committing to noisy targets and achieves better generalization, which aligns with \citet{kendall2017uncertainties} and \citet{seitzer2022pitfalls}.

\subsection{Hyperparameter Sensitivity}

\begin{table}[ht]
\centering
\small
\resizebox{\columnwidth}{!}{%
\begin{tabular}{cccccc}
\toprule
\(\alpha\) & \(\beta\) & \(\gamma\) & \textbf{NLL (Dev)} & \textbf{NLL (Test)} & \textbf{Description} \\
\midrule
1 & 1 & 1 & 1.3674 & 1.2439 & standard NLL \\
1 & 2 & 1 & 1.4718 & 1.4519 & uncertainty-averse \\
1 & 1 & 2 & 1.3363 & 1.3766 & error-focused \\
1 & 1 & 0.5 & 1.4555 & 1.4459 & calibration-first \\
\bottomrule
\end{tabular}
}
\caption{Sensitivity analysis of Gaussian NLL loss weighting parameters \(\alpha\), \(\beta\), and \(\gamma\) on the E-DAIC seq-to-seq model (single run).}
\label{tab:hyperparam_sensitivity}
\end{table}

Table~\ref{tab:hyperparam_sensitivity} reports the effect of the NLL weighting parameters \(\beta\) (log-variance term) and \(\gamma\) (normalised squared error term). The \(\alpha\) weights only the constant \(\log(2\pi)\) and therefore do not affect gradients. The standard configuration (\(\beta = \gamma = 1\)) achieves the lowest test NLL (1.2439), which indicates a balanced trade-off between data fit and uncertainty calibration. A doubled \(\beta\) (``uncertainty-averse'') raises NLL on both splits, which suggests that excessive penalisation of predicted variance compresses the model's uncertainty range. A doubled \(\gamma\) (``error-focused'') yields a marginal Dev improvement but increases the test NLL to 1.3766, consistent with overfitting. A halved \(\gamma\) (``calibration-first'') produces similarly elevated NLL, potentially because the weakened error term provides insufficient supervision for accurate mean predictions. Hence, aggressive reweighting may destabilize the balance between sharpness and calibration, and the default Gaussian NLL remains the most reliable configuration.

\begin{figure}
    \centering
    \includegraphics[width=\linewidth]{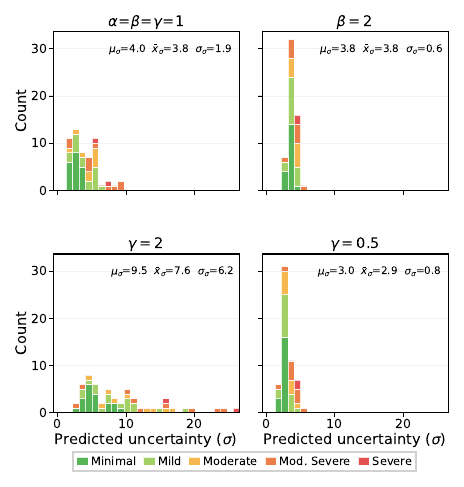}
    \caption{Predicted uncertainty (\(\sigma\)) by PHQ-8 severity category under four loss configurations: standard NLL (top-left), increased log-variance penalty \(\beta = 2\) (top-right), error-penalised \(\gamma = 2\) (bottom-left), and calibration-focused \(\gamma = 0.5\) (bottom-right). Summary statistics (\(\mu\), \(\tilde{x}\), \(\sigma\)) refer to the distribution of predicted uncertainty across all test participants.}
    \label{fig:uncertainty_dist_grouped}
\end{figure}

Beyond aggregate NLL, loss weighting also shapes how the model distributes uncertainty across severity levels (Figure~\ref{fig:uncertainty_dist_grouped}). In the standard setting with $\alpha=\beta=\gamma=1$, the predicted \(\sigma\) values span a moderate range (\(\mu_{\sigma} = 4.0\), \(\tilde{x} = 3.8\), \(\sigma_{\sigma} = 1.9\)) and the five severity categories stay within \([0, 10]\). Both the \(\beta = 2\) and \(\gamma = 0.5\) configurations compress predictions into a narrow band (\(\sigma_{\sigma} = 0.6\) and \(0.8\), respectively), which collapses severity-level distinctions and limits clinical utility. In contrast, \(\gamma = 2\) produces a wide distribution (\(\mu_{\sigma} = 9.5\), \(\sigma_{\sigma} = 6.2\)) with better separation between severity groups, yet the inflated absolute values reflect miscalibration consistent with its elevated test NLL. 


\subsection{Uncertainty Calibration and Interpretability}
\label{sec:calibration_main}

\begin{figure*}[!ht]
    \centering
    \begin{subfigure}[t]{\linewidth}
    \centering
    \includegraphics[width=\linewidth]{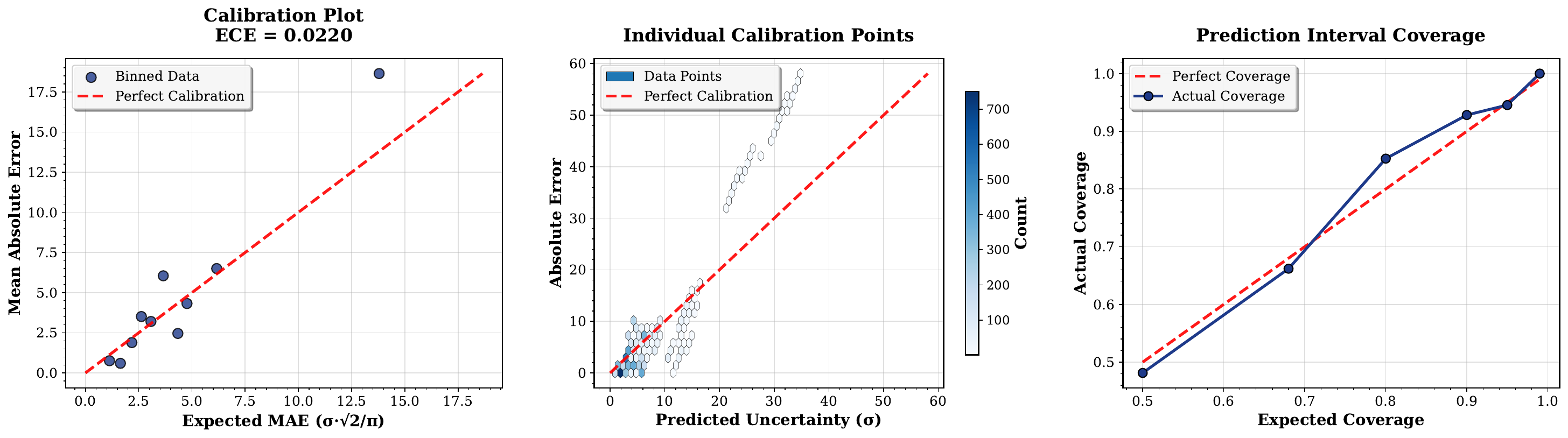}
    \caption{Probabilistic Gaussian NLL ($\alpha = \beta = \gamma = 1$)}
    \label{fig:calibration_analysis_seq-to-seq_daic_NLL}
    \end{subfigure}
    \begin{subfigure}[t]{\linewidth}
        \centering
        \includegraphics[width=\linewidth]{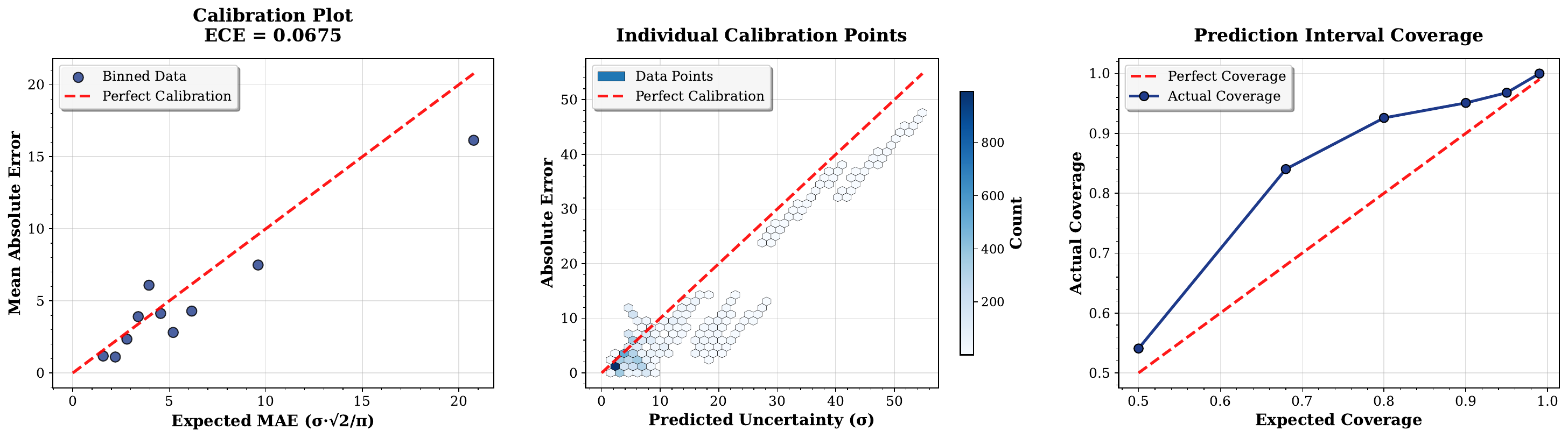}
        \caption{Point Estimate MSE}
        \label{fig:calibration_analysis_seq-to-seq_daic_MSE}
    \end{subfigure}
    \caption{Calibration analysis of PTTSD seq-to-seq on DAIC test set (Gaussian NLL vs. MSE)}
    \label{fig:calibration_analysis}
\end{figure*}

Accurate uncertainty quantification is critical in clinical NLP, where predictions may inform sensitive decisions. We first evaluate PTTSD calibration using the Expected Calibration Error (ECE), empirical coverage, and visual diagnostics in Figure~\ref{fig:calibration_analysis}. We then explore how predicted uncertainty evolves over time and correlates with error to offer insights into model interpretability and potential clinical utility.

\paragraph{Calibration Metrics.}
Figure~\ref{fig:calibration_analysis} compares models trained with Gaussian NLL and MSE losses. Each subplot presents (i) a binned calibration curve comparing predicted standard deviation and MAE; (ii) a scatter plot of predicted uncertainty vs. observed error, and (iii) a coverage plot showing the percentage of ground truth values falling within model-predicted confidence intervals. Perfect calibration aligns with the diagonal in all plots.

The Gaussian NLL model achieves a low ECE of 0.0220 and near-ideal 68\% coverage (66.2\%), indicating well-calibrated uncertainty. It adapts confidence intervals to input ambiguity, producing sharp yet reliable estimates. In contrast, the MSE-based model is underconfident, with wide intervals (84.0\% coverage) and worse calibration (ECE = 0.0675). Thus, probabilistic modeling provides more trustworthy uncertainty estimates than single-point regression.

\begin{figure}
    \centering
    \includegraphics[width=\columnwidth]{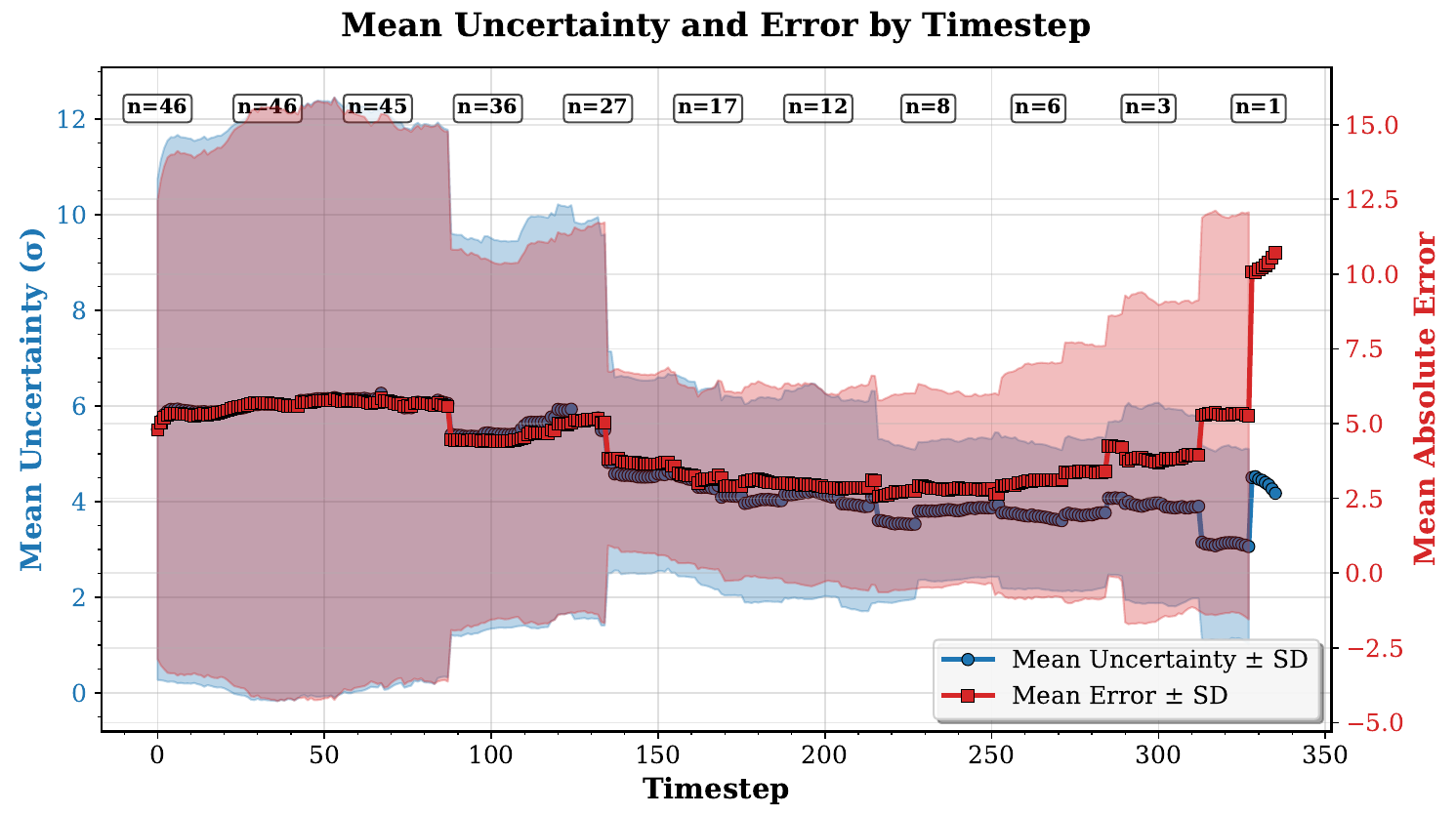}
    \caption{Temporal dynamics of uncertainty and error across utterance positions (Gaussian NLL).}
    \label{fig:temporal_tendencies}
\end{figure}

\paragraph{Interpretable Temporal Behavior.}
To understand how uncertainty evolves over a session, Figure~\ref{fig:temporal_tendencies} plots the average predicted uncertainty and absolute error across utterance positions. Initially, both metrics are high due to limited context. As the dialogue progresses, uncertainty decreases and stabilizes around timestep 250. After timestep 300, the error begins to increase again, likely due to data sparsity, since only a few sessions exceed this length in the training set.

\begin{figure}
    \centering
    \includegraphics[width=\columnwidth]{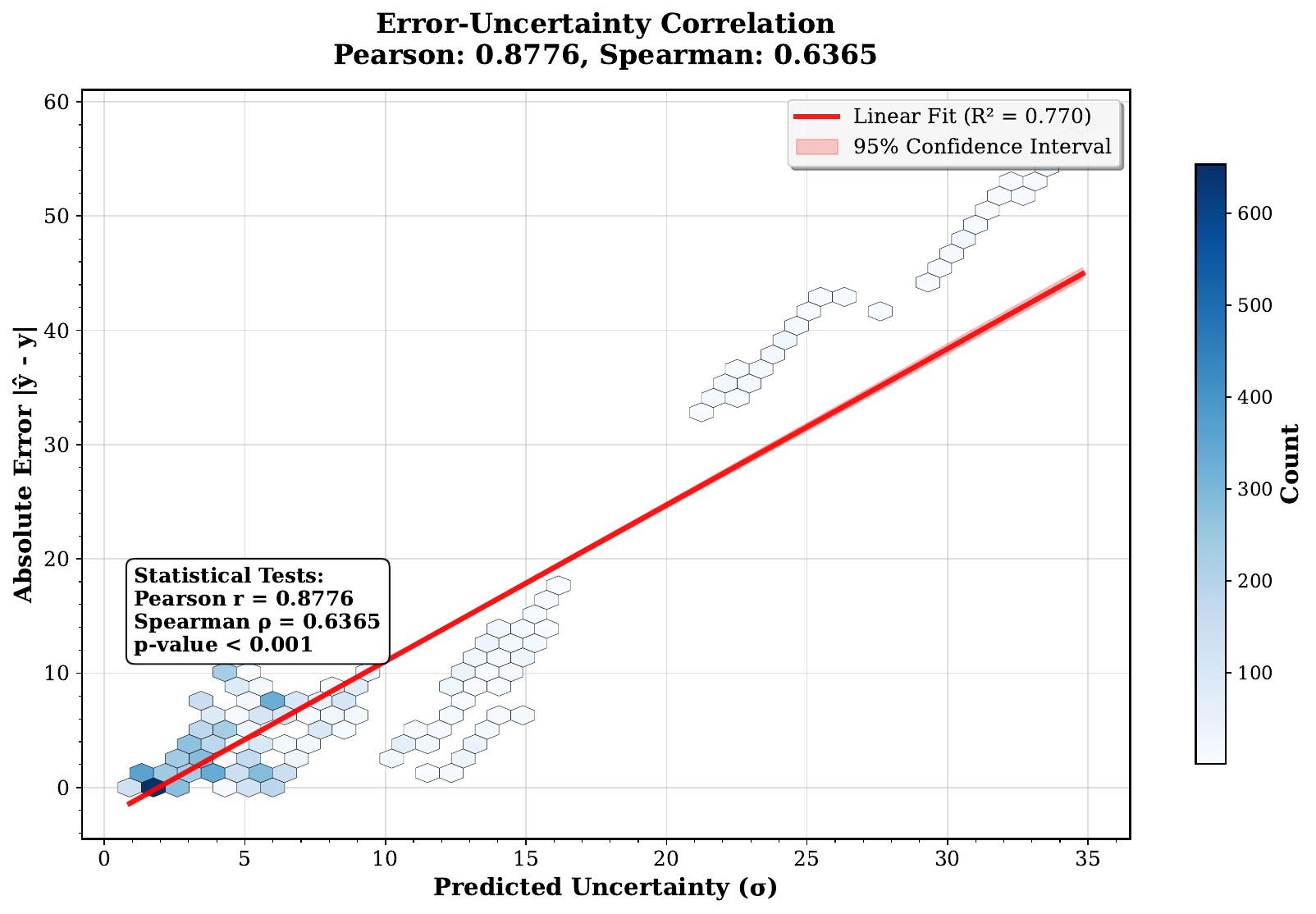}
    \caption{Correlation between predicted uncertainty and absolute error.}
    \label{fig:error_uncertainty_corr}
\end{figure}

\paragraph{Error–Uncertainty Correlation.}
In addition to aggregate calibration, we report the correlation between error and uncertainty, since a useful uncertainty signal should covary with actual prediction error so that harder inputs receive wider intervals.
As shown in Figure~\ref{fig:error_uncertainty_corr}, PTTSD exhibits a strong correlation between predicted standard deviation and absolute error ($r = 0.88$, $\rho = 0.64$, $p < 0.001$), which indicates that the model can meaningfully distinguish between confident and uncertain predictions.

\begin{figure*}[!ht]
  \centering
  \includegraphics[width=\textwidth]{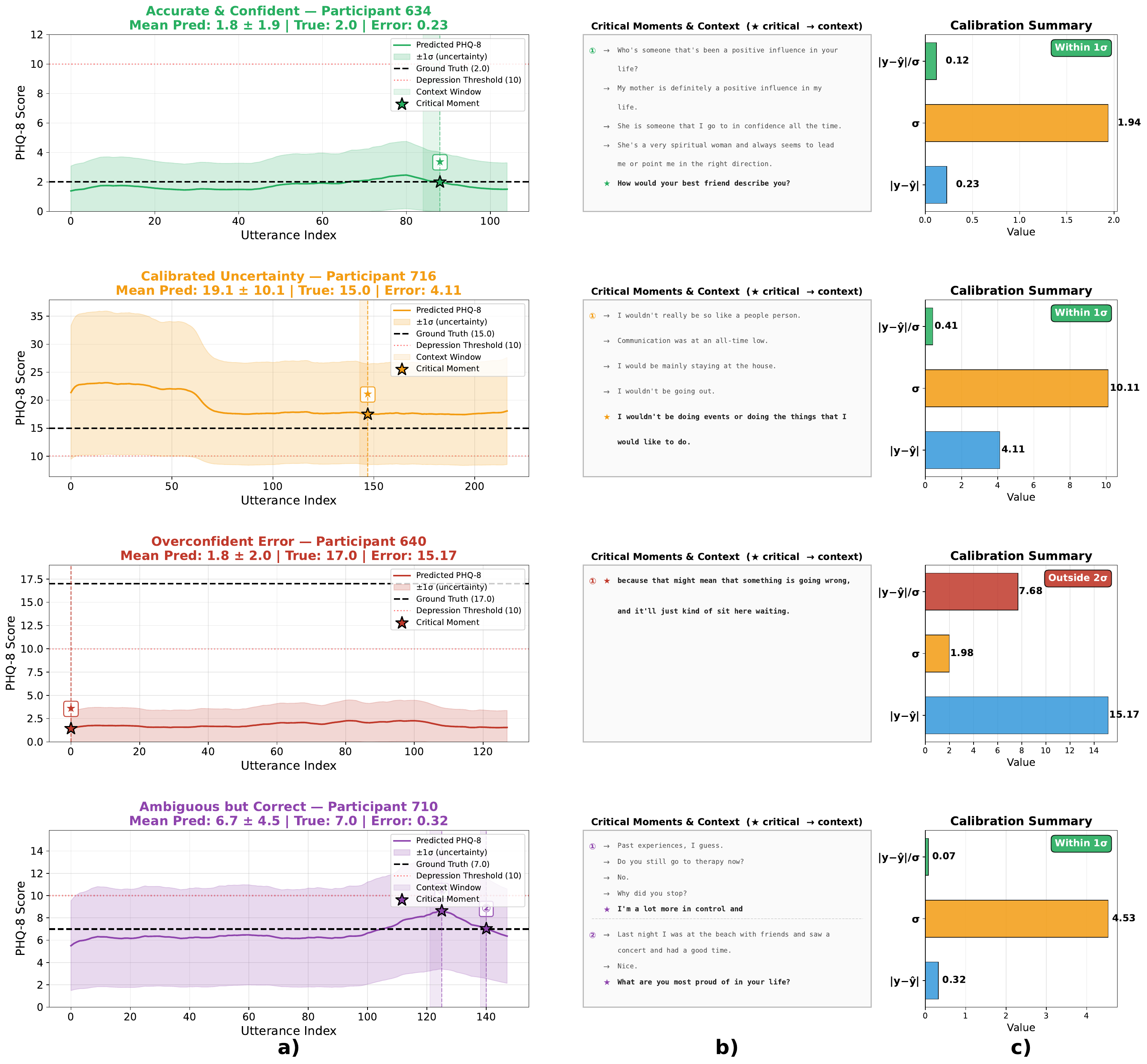}
  \caption{Case studies illustrating four uncertainty regimes. Each row shows one participant with (a)~sequence-level PHQ-8 predictions and $\pm1\sigma$ intervals, (b)~the critical context window, and (c)~a calibration summary comparing $|y - \hat{y}|$, $\sigma$, and their ratio. Rows from top to bottom: \emph{Accurate \& Confident} (Participant~634), \emph{Calibrated Uncertainty} (Participant~716), \emph{Overconfident Error} (Participant~640), and \emph{Ambiguous but Correct} (Participant~710).}
  \label{fig:case_studies}
\end{figure*}

\paragraph{Case Studies.}
Figure~\ref{fig:case_studies} illustrates sequence-level predictions for four representative participants that each capture a distinct uncertainty pattern. Participant~634 represents the ideal case, \emph{Accurate \& Confident}: the prediction trajectory remains roughly constant and closely tracks the ground truth with narrow $\pm1\sigma$ bands and a normalised error of $0.12$. Participant~716 shows \emph{Calibrated Uncertainty} during a more difficult narrative. The model initially struggles with high error until utterance 70, likely because the participant reports a lack of memorable positive experiences and a complicated relationship with their girlfriend. As the model processes more information, it converges to the ground truth, though it maintains wide confidence intervals (ratio of $0.41$), potentially due to the participant's cryptic and short responses. 

Participant~640 exposes an \emph{Overconfident Error}, the model's worst-case failure mode. Despite a true score of ~$17$, the model predicts low severity, with tight bands, resulting in a ratio of $7.68$ (outside $2\sigma$). Here, the transcript language may mask symptom severity. Finally, Participant~710 is \emph{Ambiguous but Correct}. The model picks up on mentions of sleep deprivation and a PTSD diagnosis from six years ago. The predicted PHQ-8 score reaches a peak of $\hat{y}=8.63$ at utterance 125. The peak is accompanied by increased uncertainty, likely due to a lack of signals in the preceding dialogue. Interestingly, the predicted score later decreases toward the ground truth ($|y - \hat{y}| = 0.32$, marked by \ding{73} at utterance 140) as the participant reports feeling more in control and mentions attending a concert. While the model hedges with broad intervals ($\sigma = 4.53$), it successfully captures the shifting sentiment.

Participants~716 and~640 represent extreme cases. In aggregate, however, the model's typical behavior is closer to the participants~6134 and~710 with accurate predictions and with uncertainty that reflects input ambiguity, which aligns with the quantitative evaluation presented before. Overall, the case studies visually demonstrate the practical value of uncertainty quantification. PTTSD identifies both predictions clinicians can trust and those that warrant caution.




\section{Conclusion}\label{sec:conclusion}
We introduced PTTSD, a probabilistic framework for predicting PHQ-8 depression severity from utterance-level clinical interviews. PTTSD outputs calibrated Gaussian or Student-$t$ distributions rather than point estimates, such that clinicians receive both a predicted severity and a model-confidence signal alongside it.

A key design feature is the seq-to-seq variant, which reveals how the model's belief evolves over an interview, even though the target is a single session-level score. This provides temporal interpretability as we can identify when the model becomes confident, which utterances drive prediction shifts, and where ambiguity persists. Our calibration analyses confirm that predicted uncertainty is well-aligned with actual error ($r = 0.88$, ECE = 0.022), and case studies illustrate how these properties manifest at the individual participant level.

Our experiments on DAIC and E-DAIC show competitive performance among text-only systems. While accuracy degrades on the original E-DAIC transcripts relative to WhisperX re-transcriptions, the calibration and uncertainty properties — our main contribution — hold across transcript sources.
Future work includes multimodal extensions to complement text with prosodic and visual signals, and clinical validation to evaluate whether uncertainty estimates inform practitioner decisions.

\section*{Limitations}

While PTTSD offers promising results in predictive accuracy and uncertainty modeling, several limitations remain. First, the framework relies solely on textual data. Although effective, it does not leverage multimodal cues such as vocal prosody or facial expressions, which are known to be informative for assessing mental health. Second, the E-DAIC dataset contains fewer than 300 participants, and further reduction due to filtering and partitioning limits the statistical power and generalizability of our findings to broader clinical settings. Third, the interviews in E-DAIC are conducted with a virtual interviewer ("Ellie") operated in a Wizard-of-Oz setup rather than a real clinician, which may affect the ecological validity of the speech data and limit applicability to authentic client–clinician interactions. In terms of modeling, we encode utterances independently using pretrained language models without context-aware finetuning, potentially overlooking local coherence or discourse-level cues. Furthermore, while PTTSD provides distributional predictions, we do not assess its clinical utility or decision-support value. Human-centered evaluations with therapists or end users are needed to determine the interpretability and trustworthiness of predicted uncertainty. Finally, although we evaluate calibration quantitatively, we do not study how uncertainty scores might be perceived or utilized by clinicians in real-world settings.


\bibliography{anthology,custom}

\begin{thebibliography}{37}
\providecommand{\natexlab}[1]{#1}

\bibitem[{Al~Hanai et~al.(2018)Al~Hanai, Ghassemi, and
  Glass}]{alhanai2018sequence}
Tuka Al~Hanai, Mohammad~M Ghassemi, and James~R Glass. 2018.
\newblock Detecting depression with audio/text sequence modeling of interviews.
\newblock In \emph{Interspeech}, pages 1716--1720.

\bibitem[{Burdisso et~al.(2023)Burdisso, Villatoro-Tello, Madikeri, and
  Motlicek}]{burdisso2023graph}
Sergio Burdisso, Esaú Villatoro-Tello, Srikanth Madikeri, and Petr Motlicek.
  2023.
\newblock \href {https://doi.org/10.21437/Interspeech.2023-1923} {Node-weighted
  graph convolutional network for depression detection in transcribed clinical
  interviews}.
\newblock In \emph{INTERSPEECH 2023}, pages 3617--3621.

\bibitem[{Chen et~al.(2024)Chen, Deng, Zhou, Wu, Qian, and
  Huang}]{chen-etal-2024-depression}
Zhuang Chen, Jiawen Deng, Jinfeng Zhou, Jincenzi Wu, Tieyun Qian, and Minlie
  Huang. 2024.
\newblock \href {https://doi.org/10.18653/v1/2024.naacl-long.452} {Depression
  detection in clinical interviews with {LLM}-empowered structural element
  graph}.
\newblock In \emph{Proceedings of the 2024 Conference of the North American
  Chapter of the Association for Computational Linguistics: Human Language
  Technologies (Volume 1: Long Papers)}, pages 8181--8194, Mexico City, Mexico.
  Association for Computational Linguistics.

\bibitem[{{DAIC‑WOZ Project}(2019)}]{edaic2020}
{DAIC‑WOZ Project}. 2019.
\newblock \href {https://dcapswoz.ict.usc.edu/extended-daic-database-download/}
  {Extended distress analysis interview corpus–wizard of oz (e‑daic)}.
\newblock Extended DAIC Database, downloadable via the DAIC‑WOZ project
  website at dcapswoz.ict.usc.edu.
\newblock AVEC 2019 subset: 275 sessions (163 train, 56 dev, 56 test);
  includes audio, transcripts, visual and acoustic features; Accessed:
  2025-01-30.

\bibitem[{Dia et~al.(2024)Dia, Khodabandelou, and Othmani}]{dia2024paying}
Mamadou Dia, Ghazaleh Khodabandelou, and Alice Othmani. 2024.
\newblock Paying attention to uncertainty: A stochastic multimodal transformers
  for post-traumatic stress disorder detection using video.
\newblock \emph{Computer Methods and Programs in Biomedicine}, 257:108439.

\bibitem[{Fang et~al.(2023{\natexlab{a}})Fang, Peng, Liang, Hung, and
  Liu}]{fang2023multimodal}
Ming Fang, Siyu Peng, Yujia Liang, Chih-Cheng Hung, and Shuhua Liu.
  2023{\natexlab{a}}.
\newblock A multimodal fusion model with multi-level attention mechanism for
  depression detection.
\newblock \emph{Biomedical Signal Processing and Control}, 82:104561.

\bibitem[{Fang et~al.(2023{\natexlab{b}})Fang, Peng, Liang, Hung, and
  Liu}]{fang2023transformer}
Ming Fang, Siyu Peng, Yujia Liang, Chih-Cheng Hung, and Shuhua Liu.
  2023{\natexlab{b}}.
\newblock A multimodal fusion model with multi-level attention mechanism for
  depression detection.
\newblock \emph{Biomedical Signal Processing and Control}, 82:104561.

\bibitem[{Gong and Poellabauer(2017)}]{gong2020topic}
Yuan Gong and Christian Poellabauer. 2017.
\newblock \href {https://doi.org/10.1145/3133944.3133945} {Topic modeling based
  multi-modal depression detection}.
\newblock In \emph{Proceedings of the 7th Annual Workshop on Audio/Visual
  Emotion Challenge}, AVEC '17, page 69–76, New York, NY, USA. Association
  for Computing Machinery.

\bibitem[{Gratch et~al.(2014)Gratch, Artstein, Lucas, Stratou, Scherer,
  Nazarian, Wood, Boberg, DeVault, Marsella, Traum, Rizzo, and
  Morency}]{gratch-etal-2014-distress}
Jonathan Gratch, Ron Artstein, Gale Lucas, Giota Stratou, Stefan Scherer,
  Angela Nazarian, Rachel Wood, Jill Boberg, David DeVault, Stacy Marsella,
  David Traum, Skip Rizzo, and Louis-Philippe Morency. 2014.
\newblock \href {https://aclanthology.org/L14-1421/} {The distress analysis
  interview corpus of human and computer interviews}.
\newblock In \emph{Proceedings of the Ninth International Conference on
  Language Resources and Evaluation ({LREC}`14)}, pages 3123--3128, Reykjavik,
  Iceland. European Language Resources Association (ELRA).

\bibitem[{Guo et~al.(2022)Guo, Zhu, Hao, and Hong}]{guo2022topic}
Yanrong Guo, Chenyang Zhu, Shijie Hao, and Richang Hong. 2022.
\newblock A topic-attentive transformer-based model for multimodal depression
  detection.
\newblock \emph{arXiv preprint arXiv:2206.13256}.

\bibitem[{Ji et~al.(2022)Ji, Zhang, Ansari, Fu, Tiwari, and
  Cambria}]{ji-etal-2022-mentalbert}
Shaoxiong Ji, Tianlin Zhang, Luna Ansari, Jie Fu, Prayag Tiwari, and Erik
  Cambria. 2022.
\newblock \href {https://aclanthology.org/2022.lrec-1.778/} {{M}ental{BERT}:
  Publicly available pretrained language models for mental healthcare}.
\newblock In \emph{Proceedings of the Thirteenth Language Resources and
  Evaluation Conference}, pages 7184--7190, Marseille, France. European
  Language Resources Association.

\bibitem[{Kendall and Gal(2017)}]{kendall2017uncertainties}
Alex Kendall and Yarin Gal. 2017.
\newblock \href
  {https://proceedings.neurips.cc/paper_files/paper/2017/file/2650d6089a6d640c5e85b2b88265dc2b-Paper.pdf}
  {What uncertainties do we need in bayesian deep learning for computer
  vision?}
\newblock In \emph{Advances in Neural Information Processing Systems},
  volume~30. Curran Associates, Inc.

\bibitem[{Kroenke et~al.(2009)Kroenke, Strine, Spitzer, Williams, Berry, and
  Mokdad}]{kroenke2009phq}
Kurt Kroenke, Tara~W Strine, Robert~L Spitzer, Janet~BW Williams, Joyce~T
  Berry, and Ali~H Mokdad. 2009.
\newblock The phq-8 as a measure of current depression in the general
  population.
\newblock \emph{Journal of affective disorders}, 114(1-3):163--173.

\bibitem[{Li and Zhou(2025)}]{li2025fair}
Yonghong Li and Xiuzhuang Zhou. 2025.
\newblock Fair uncertainty quantification for depression prediction.
\newblock \emph{arXiv preprint arXiv:2505.04931}.

\bibitem[{Makiuchi et~al.(2019)Makiuchi, Warnita, Uto, and
  Shinoda}]{makiuchi2019multimodal}
Mariana~Rodrigues Makiuchi, Tifani Warnita, Kuniaki Uto, and Koichi Shinoda.
  2019.
\newblock \href {https://doi.org/10.1145/3347320.3357694} {Multimodal fusion of
  bert-cnn and gated cnn representations for depression detection}.
\newblock In \emph{Proceedings of the 9th International on Audio/Visual Emotion
  Challenge and Workshop}, AVEC '19, page 55–63, New York, NY, USA.
  Association for Computing Machinery.

\bibitem[{Mandal et~al.(2025)Mandal, Atzil-Slonim, Solorio, and
  Gurevych}]{mandal-etal-2025-enhancing}
Aishik Mandal, Dana Atzil-Slonim, Thamar Solorio, and Iryna Gurevych. 2025.
\newblock \href {https://aclanthology.org/2025.clpsych-1.4/} {Enhancing
  depression detection via question-wise modality fusion}.
\newblock In \emph{Proceedings of the 10th Workshop on Computational
  Linguistics and Clinical Psychology (CLPsych 2025)}, pages 44--61,
  Albuquerque, New Mexico. Association for Computational Linguistics.

\bibitem[{Mao et~al.(2023)Mao, Zhang, Wang, Li, Jiao, Zhu, Wu, Zheng, Qian,
  Lyu, Ye, and Chen}]{mao2022prediction}
Kaining Mao, Wei Zhang, Deborah~Baofeng Wang, Ang Li, Rongqi Jiao, Yanhui Zhu,
  Bin Wu, Tiansheng Zheng, Lei Qian, Wei Lyu, Minjie Ye, and Jie Chen. 2023.
\newblock \href {https://doi.org/10.1109/TAFFC.2022.3154332} {Prediction of
  depression severity based on the prosodic and semantic features with
  bidirectional lstm and time distributed cnn}.
\newblock \emph{IEEE Transactions on Affective Computing}, 14(3):2251--2265.

\bibitem[{Muennighoff et~al.(2023)Muennighoff, Tazi, Magne, and
  Reimers}]{muennighoff-etal-2023-mteb}
Niklas Muennighoff, Nouamane Tazi, Loic Magne, and Nils Reimers. 2023.
\newblock \href {https://doi.org/10.18653/v1/2023.eacl-main.148} {{MTEB}:
  Massive text embedding benchmark}.
\newblock In \emph{Proceedings of the 17th Conference of the European Chapter
  of the Association for Computational Linguistics}, pages 2014--2037,
  Dubrovnik, Croatia. Association for Computational Linguistics.

\bibitem[{Niu et~al.(2021)Niu, Chen, Chen, and Yang}]{niu2019hierarchical}
Meng Niu, Kai Chen, Qingcai Chen, and Lufeng Yang. 2021.
\newblock \href {https://doi.org/10.1109/ICASSP39728.2021.9413486} {Hcag: A
  hierarchical context-aware graph attention model for depression detection}.
\newblock In \emph{ICASSP 2021 - 2021 IEEE International Conference on
  Acoustics, Speech and Signal Processing (ICASSP)}, pages 4235--4239.

\bibitem[{Nykoniuk et~al.(2025)Nykoniuk, Basystiuk, Shakhovska, and
  Melnykova}]{nykoniuk2025multimodal}
Mariia Nykoniuk, Oleh Basystiuk, Nataliya Shakhovska, and Nataliia Melnykova.
  2025.
\newblock Multimodal data fusion for depression detection approach.
\newblock \emph{Computation}, 13(1):9.

\bibitem[{Oureshi et~al.(2021)Oureshi, Dias, Saha, and
  Hasanuzzaman}]{oureshi2017deep}
Syed~Arbaaz Oureshi, Gaël Dias, Sriparna Saha, and Mohammed Hasanuzzaman.
  2021.
\newblock \href {https://doi.org/10.1109/IJCNN52387.2021.9534330} {Gender-aware
  estimation of depression severity level in a multimodal setting}.
\newblock In \emph{2021 International Joint Conference on Neural Networks
  (IJCNN)}, pages 1--8.

\bibitem[{Paszke et~al.(2019)Paszke, Gross, Massa, Lerer, Bradbury, Chanan,
  Killeen, Lin, Gimelshein, Antiga, Desmaison, Kopf, Yang, DeVito, Raison,
  Tejani, Chilamkurthy, Steiner, Fang, Bai, and Chintala}]{paszke2019pytorch}
Adam Paszke, Sam Gross, Francisco Massa, Adam Lerer, James Bradbury, Gregory
  Chanan, Trevor Killeen, Zeming Lin, Natalia Gimelshein, Luca Antiga, Alban
  Desmaison, Andreas Kopf, Edward Yang, Zachary DeVito, Martin Raison, Alykhan
  Tejani, Sasank Chilamkurthy, Benoit Steiner, Lu~Fang, and 2 others. 2019.
\newblock \href
  {https://papers.nips.cc/paper_files/paper/2019/hash/bdbca288fee7f92f2bfa9f7012727740-Abstract.html}
  {Pytorch: An imperative style, high-performance deep learning library}.
\newblock In \emph{Advances in Neural Information Processing Systems},
  volume~32.

\bibitem[{Qureshi et~al.(2019{\natexlab{a}})Qureshi, Hasanuzzaman, Saha, and
  Dias}]{qureshi2017ensemble}
Syed~Arbaaz Qureshi, Mohammed Hasanuzzaman, Sriparna Saha, and Ga{\"e}l Dias.
  2019{\natexlab{a}}.
\newblock The verbal and non verbal signals of depression--combining acoustics,
  text and visuals for estimating depression level.
\newblock \emph{arXiv preprint arXiv:1904.07656}.

\bibitem[{Qureshi et~al.(2019{\natexlab{b}})Qureshi, Saha, Hasanuzzaman, and
  Dias}]{qureshi2019multitask}
Syed~Arbaaz Qureshi, Sriparna Saha, Mohammed Hasanuzzaman, and Ga{\"e}l Dias.
  2019{\natexlab{b}}.
\newblock Multitask representation learning for multimodal estimation of
  depression level.
\newblock \emph{IEEE Intelligent Systems}, 34(5):45--52.

\bibitem[{Ray et~al.(2019)Ray, Kumar, Reddy, Mukherjee, and
  Garg}]{ray2019multilevel}
Anupama Ray, Siddharth Kumar, Rutvik Reddy, Prerana Mukherjee, and Ritu Garg.
  2019.
\newblock \href {https://doi.org/10.1145/3347320.3357697} {Multi-level
  attention network using text, audio and video for depression prediction}.
\newblock In \emph{Proceedings of the 9th International on Audio/Visual Emotion
  Challenge and Workshop}, AVEC '19, page 81–88, New York, NY, USA.
  Association for Computing Machinery.

\bibitem[{Reimers and Gurevych(2019)}]{reimers-gurevych-2019-sentence}
Nils Reimers and Iryna Gurevych. 2019.
\newblock \href {https://doi.org/10.18653/v1/D19-1410} {Sentence-{BERT}:
  Sentence embeddings using {S}iamese {BERT}-networks}.
\newblock In \emph{Proceedings of the 2019 Conference on Empirical Methods in
  Natural Language Processing and the 9th International Joint Conference on
  Natural Language Processing (EMNLP-IJCNLP)}, pages 3982--3992, Hong Kong,
  China. Association for Computational Linguistics.

\bibitem[{Rohanian et~al.(2019)Rohanian, Hough, and
  Purver}]{rohanian2019contextual}
Morteza Rohanian, Julian Hough, and Matthew Purver. 2019.
\newblock \href {https://doi.org/10.21437/Interspeech.2019-2283} {Detecting
  depression with word-level multimodal fusion}.
\newblock In \emph{INTERSPEECH 2019}, pages 1443--1447.

\bibitem[{Sadeghi et~al.(2023)Sadeghi, Egger, Agahi, Richer, Capito, Rupp,
  Schindler-Gmelch, Berking, and Eskofier}]{sadeghi2023depression}
Misha Sadeghi, Bernhard Egger, Reza Agahi, Robert Richer, Klara Capito,
  Lydia~Helene Rupp, Lena Schindler-Gmelch, Matthias Berking, and Bjoern~M.
  Eskofier. 2023.
\newblock \href {https://doi.org/10.1109/BHI58575.2023.10313367} {Exploring the
  capabilities of a language model-only approach for depression detection in
  text data}.
\newblock In \emph{2023 IEEE EMBS International Conference on Biomedical and
  Health Informatics (BHI)}, pages 1--5.

\bibitem[{Sadeghi et~al.(2024)Sadeghi, Richer, Egger, Schindler-Gmelch, Rupp,
  Rahimi, Berking, and Eskofier}]{sadeghi2024harnessing}
Misha Sadeghi, Robert Richer, Bernhard Egger, Lena Schindler-Gmelch,
  Lydia~Helene Rupp, Farnaz Rahimi, Matthias Berking, and Bjoern~M Eskofier.
  2024.
\newblock Harnessing multimodal approaches for depression detection using large
  language models and facial expressions.
\newblock \emph{npj Mental Health Research}, 3(1):66.

\bibitem[{Seitzer et~al.(2022)Seitzer, Tavakoli, Antic, and
  Martius}]{seitzer2022pitfalls}
Maximilian Seitzer, Arash Tavakoli, Dimitrije Antic, and Georg Martius. 2022.
\newblock \href {https://openreview.net/forum?id=aPOpXlnV1T} {On the pitfalls
  of heteroscedastic uncertainty estimation with probabilistic neural
  networks}.
\newblock In \emph{International Conference on Learning Representations}.

\bibitem[{Stepanov et~al.(2018)Stepanov, Lathuilière, Chowdhury, Ghosh,
  Vieriu, Sebe, and Riccardi}]{stepanov2021multimodal}
Evgeny~A. Stepanov, Stéphane Lathuilière, Shammur~Absar Chowdhury, Arindam
  Ghosh, Radu-Laurenţiu Vieriu, Nicu Sebe, and Giuseppe Riccardi. 2018.
\newblock \href {https://doi.org/10.1109/HealthCom.2018.8531119} {Depression
  severity estimation from multiple modalities}.
\newblock In \emph{2018 IEEE 20th International Conference on e-Health
  Networking, Applications and Services (Healthcom)}, pages 1--6.

\bibitem[{WHO(2017)}]{who2017depression}
WHO. 2017.
\newblock \href
  {https://www.who.int/publications/i/item/depression-global-health-estimates}
  {Depression and other common mental disorders: Global health estimates}.
\newblock Technical report, World Health Organization, Geneva.
\newblock WHO/MSD/MER/2017.2.

\bibitem[{WHO(2022)}]{who2022mentalhealth}
WHO. 2022.
\newblock \href {https://www.who.int/publications/i/item/9789240049338} {World
  mental health report: Transforming mental health for all}.
\newblock Accessed: 2025-05-18.

\bibitem[{Williamson et~al.(2016)Williamson, Godoy, Cha, Schwarzentruber,
  Khorrami, Gwon, Kung, Dagli, and Quatieri}]{williamson2016speech}
James~R. Williamson, Elizabeth Godoy, Miriam Cha, Adrianne Schwarzentruber,
  Pooya Khorrami, Youngjune Gwon, Hsiang-Tsung Kung, Charlie Dagli, and
  Thomas~F. Quatieri. 2016.
\newblock \href {https://doi.org/10.1145/2988257.2988263} {Detecting depression
  using vocal, facial and semantic communication cues}.
\newblock In \emph{Proceedings of the 6th International Workshop on
  Audio/Visual Emotion Challenge}, AVEC '16, page 11–18, New York, NY, USA.
  Association for Computing Machinery.

\bibitem[{Yang et~al.(2017)Yang, Jiang, Xia, Pei, Oveneke, and
  Sahli}]{yang2016context}
Le~Yang, Dongmei Jiang, Xiaohan Xia, Ercheng Pei, Meshia~C\'{e}dric Oveneke,
  and Hichem Sahli. 2017.
\newblock \href {https://doi.org/10.1145/3133944.3133948} {Multimodal
  measurement of depression using deep learning models}.
\newblock In \emph{Proceedings of the 7th Annual Workshop on Audio/Visual
  Emotion Challenge}, AVEC '17, page 53–59, New York, NY, USA. Association
  for Computing Machinery.

\bibitem[{Zhang and Guo(2024)}]{zhang2024prompt}
Jun Zhang and Yanrong Guo. 2024.
\newblock \href {https://doi.org/10.1016/j.patrec.2024.01.005} {Multilevel
  depression status detection based on fine-grained prompt learning}.
\newblock \emph{Pattern Recogn. Lett.}, 178(C):167–173.

\bibitem[{Zhang et~al.(2025)Zhang, Li, Chen, Zheng, and Li}]{zhang2025mil}
Xu~Zhang, Chenlong Li, Weisi Chen, Jiaxin Zheng, and Feihong Li. 2025.
\newblock Optimizing depression detection in clinical doctor-patient interviews
  using a multi-instance learning framework.
\newblock \emph{Scientific Reports}, 15(1):6637.

\end{thebibliography}

\appendix

\section{PHQ-8 Depression Assessment}\label{app:phq8}

The Patient Health Questionnaire-8 (PHQ-8) \cite{kroenke2009phq} is a widely used self-report scale designed to measure the presence and severity of depressive symptoms. It is derived from the PHQ-9 but omits the ninth item concerning suicidal thoughts, making it more suitable for large-scale screening and automated processing.

Each of the eight items corresponds to a DSM-IV criterion for depression and asks respondents to rate how often they have experienced a specific symptom over the past two weeks. Responses are scored on a 4-point Likert scale:
\begin{itemize}
    \item 0 – Not at all
    \item 1 – Several days
    \item 2 – More than half the days
    \item 3 – Nearly every day
\end{itemize}

The total PHQ-8 score ranges from 0 to 24 and is interpreted as follows:
\begin{itemize}
    \item 0–4: None
    \item 5–9: Mild depression
    \item 10–14: Moderate depression
    \item 15–19: Moderately severe depression
    \item 20–24: Severe depression
\end{itemize}

The PHQ-8 has been validated in both clinical and general populations and is considered a reliable proxy for identifying depressive symptom severity in mental health research.

\section{Distress Analysis Interview Corpus (DAIC and E-DAIC)}\label{app:edaic}

The \textbf{Distress Analysis Interview Corpus} (DAIC-WOZ)~\cite{gratch-etal-2014-distress} and its extended version, \textbf{E-DAIC}~\cite{edaic2020}, are widely used datasets for research in automated depression detection. Both datasets contain semi-structured clinical interviews conducted by a virtual interviewer named Ellie, operated via a "Wizard-of-Oz" setup, to elicit verbal and non-verbal indicators of psychological distress.

\subsection{E-DAIC vs. DAIC-WOZ}

The E-DAIC corpus is a re-transcribed and quality-controlled extension of DAIC-WOZ. It corrects known transcription errors and inconsistencies, and provides standardized splits for training, development, and testing. While DAIC-WOZ has been extensively used in prior work, E-DAIC offers improved data quality and is recommended for text-based modeling tasks.

\subsection{Dataset Composition}

E-DAIC consists of 275 participant interviews, partitioned as follows:

\begin{itemize}
    \item \textbf{Training set}: 163 participants
    \item \textbf{Development set}: 56 participants
    \item \textbf{Test set}: 56 participants
\end{itemize}

Each session includes:

\begin{itemize}
    \item \textbf{Audio recordings}: Interview audio in WAV format.
    \item \textbf{Transcripts}: Time-stamped dialogue with speaker labels.
    \item \textbf{Visual features}: Extracted using OpenFace, including facial landmarks, action units, and head pose.
    \item \textbf{Acoustic features}: Extracted via COVAREP and FORMANT analysis.
    \item \textbf{PHQ-8 scores}: Self-reported ratings of depression severity.
\end{itemize}

\subsection{Data Organization}

The dataset is organized into session-specific folders identified by participant IDs (e.g., \texttt{300\_P}), each containing:

\begin{itemize}
    \item \texttt{TRANSCRIPT.csv}: Annotated dialogue transcript.
    \item \texttt{AUDIO.wav}: Raw audio file.
    \item \texttt{COVAREP.csv}, \texttt{FORMANT.csv}: Acoustic features.
    \item \texttt{CLNF\_features.txt}, \texttt{CLNF\_AUs.csv}, \texttt{CLNF\_pose.txt}, \texttt{CLNF\_gaze.txt}: Visual features extracted using OpenFace.
\end{itemize}

Additional metadata includes:

\begin{itemize}
    \item \texttt{train\_split.csv}, \texttt{dev\_split.csv}, \texttt{test\_split.csv}: Partition definitions.
    \item \texttt{PHQ8\_scores.csv}: Item-level and total PHQ-8 responses.
\end{itemize}

\subsection{PHQ-8 Score Distribution}

PHQ-8 scores in both DAIC and E-DAIC range from 0 to 24, capturing varying levels of depressive symptom severity. The distribution is right-skewed, with a concentration of low-to-moderate severity cases, which presents challenges for model calibration and minority class performance.

\subsection{Usage Considerations}

Researchers working with DAIC or E-DAIC should consider the following:

\begin{itemize}
    \item \textbf{Data Quality}: E-DAIC addresses known issues in DAIC-WOZ, including transcript errors and missing data, and is recommended for textual modeling.
    \item \textbf{Ethical Use}: Given the sensitive nature of the interviews, ethical guidelines and approvals must be followed.
    \item \textbf{Licensing}: Access requires agreement to the dataset’s End User License Agreement (EULA).
\end{itemize}

Our use of both datasets complies with their intended research purpose. The corpora were released to support research on automated detection of psychological distress and related mental health conditions. In this work, we focus exclusively on the prediction of PHQ-8 depression severity from textual transcripts, a primary task for which the dataset was designed. The datasets are anonymized at source, with personally identifiable information removed prior to distribution. We further restrict our usage to non-commercial, academic settings, operate solely on de-identified utterance sequences, and report only aggregate results. No individual-level data or metadata are released. All use complies with the dataset’s End User License Agreement (EULA) and contributes to its intended goal of advancing computational methods for mental health assessment.

For detailed information on data preprocessing and feature extraction methodologies, refer to the official documentation provided with the dataset.

\section{Ablation Study Experimental Setup}\label{app:ablation}
For each ablation, we use the same data splits, batch size, optimizer, learning rate schedule, and early stopping criteria as the main experiments. The following configurations are evaluated:
\begin{itemize}
    \item \textbf{Full Model}: All components enabled (attention, residual, variance).
    \item \textbf{No Attention}: Attention layer removed.
    \item \textbf{No Residual}: Residual connection removed.
    \item \textbf{No Variance}: Variance prediction head disabled; model trained with MSE loss.
\end{itemize}
Each model is trained for the same number of epochs with fixed random seeds for reproducibility. After training, we evaluate on the held-out test set and report MAE, RMSE, and NLL (where available). All code, configurations, and results are available for reproducibility.


\section{Implementation Details}
\label{app:impl}

\subsection{Implementation.}
All models are implemented in PyTorch \cite{paszke2019pytorch}. Padding, batching, and masking ensure that variable-length sequences do not affect loss or metric computations.

\subsection{Hardware.}
Training is performed on a single NVIDIA A100-SXM4-80GB GPU with 80GB of GDDR6 VRAM, using CUDA version 12.2.

\subsection{Runtime.}
Training PTTSD for 50 epochs on a single NVIDIA A100–80 GB takes
\textasciitilde2h 23min in wall-clock time
(\(\approx\)172 s per epoch). The model has a total 2,703,403 trainable parameters.

\subsection{Terminology.}
Throughout, we avoid “valid/invalid utterances.” We instead say \emph{padded positions are masked} and we compute losses/metrics over \emph{non-padded} positions only.

\subsection{Batching, Padding, and Masking}
\label{app:batching}
We batch at the \emph{participant/session} level. Variable-length sequences are right-padded to the maximum length in the batch. A Boolean mask $\mathbf{m}\!\in\!\{0,1\}^{T}$ (per sequence) is propagated so that:
(i) attention, (ii) pooling, (iii) loss, and (iv) metric computations exclude padded positions.
This mask is applied within the attention mechanism and used to zero-out contributions from padded indices.

\subsection{Pooling Mechanics (seq-to-one)}
\label{app:pooling}
For the seq-to-one variant, we apply \emph{average pooling over time} on the attended sequence $\mathbf{A}\!\in\!\mathbb{R}^{T\times H}$:
\[
\bar{\mathbf{a}} \;=\; \frac{1}{\sum_t m_t} \sum_{t=1}^T m_t\, \mathbf{a}_t,
\]
where $m_t\!\in\!\{0,1\}$ masks out padding. (In the main text, we simply refer to this as \emph{average pooling}; masking only excludes padding and does not introduce a new modeling component.)

\subsection{Tokenization and Utterance Embeddings}
\label{app:embeddings}
We evaluate two encoders:
(i) \texttt{all-MiniLM-L6-v2} (Sentence-Transformers) with mean pooling over tokens; and
(ii) \texttt{MentalBERT} with the final \texttt{[CLS]} vector as utterance embedding.
We follow each model’s default casing, tokenization, and truncation rules. The stacked utterance matrix is $\mathbf{X}\!\in\!\mathbb{R}^{T\times D}$ (or $B\times T\times D$ in batched form).

\subsection{Optimization, Schedules, and Targets}
\label{app:opt}
We use Adam with cosine-annealed learning rate, training for 50 epochs with early stopping on development MAE (patience 15).
Initial learning rate $2\times10^{-4}$ decays smoothly to $10^{-4}$.
We apply a log-transform to targets during training for stability and invert it at evaluation.

\section{Predictive Distributions and NLL Details}
\label{app:nll}

\subsection{Loss Aggregation: seq-to-one vs.\ seq-to-seq}
\label{app:agg}
For \textbf{seq-to-one}, the loss is the negative log-likelihood (NLL) of the session-level prediction:
\[
\mathcal{L}_{\text{seq-to-one}} \;=\; -\log p\!\left(y \mid e_{1:T}; \theta\right).
\]
For \textbf{seq-to-seq}, we average the per-step NLL across non-padded time steps:
\[
\mathcal{L}_{\text{seq-to-seq}} \;=\; -\frac{1}{\sum_t m_t}\sum_{t=1}^{T} m_t \,\log p\!\left(y \mid e_{\le t}; \theta\right).
\]

\subsection{Gaussian Negative Log-Likelihood}
\label{app:gauss}
With predicted mean $\hat{\mu}_t$ and standard deviation $\hat{\sigma}_t$,
\begin{multline*}
\mathcal{L}_{\text{Gauss}} = \frac{1}{M}\sum_{t=1}^{T} m_t \Bigl[
  \alpha\log(2\pi) + \beta \log \hat{\sigma}_t^{2} \\
  + \gamma\,(y-\hat{\mu}_t)^2 / \hat{\sigma}_t^{2}
\Bigr].
\end{multline*}
Unless stated otherwise, $\alpha\!=\!\beta\!=\!\gamma\!=\!1$.

\subsection{Student’s-$t$ Density}
\label{app:t_density}
With $(\hat{\mu}_t,\hat{\sigma}_t,\nu_t)$,
\[
p(y)=\frac{\Gamma\!\left(\frac{\nu+1}{2}\right)}
{\Gamma\!\left(\frac{\nu}{2}\right)\sqrt{\nu\pi}\,\hat{\sigma}}
\left[1+\frac{1}{\nu}\left(\frac{y-\hat{\mu}}{\hat{\sigma}}\right)^2\right]^{-\frac{\nu+1}{2}}.
\]
We observed that Gaussian heads were the most stable and best calibrated in our setting; Student’s-$t$ is included for completeness.

\subsection{Auxiliary Objectives}
\label{app:aux}
We report MAE/MSE baselines for reference:
\[
\mathcal{L}_{\text{MSE}}=\frac{1}{\sum_t m_t}\sum_{t=1}^{T} m_t\,(y-\hat{\mu}_t)^2,
\]
\[
\mathcal{L}_{\text{MAE}}=\frac{1}{\sum_t m_t}\sum_{t=1}^{T} m_t\,|y-\hat{\mu}_t|.
\]

\section{Results on Original E-DAIC Transcripts}
\label{app:original_transcripts}

To ensure fair comparison with prior work that used the original E-DAIC transcripts rather than WhisperX re-transcriptions, we evaluate PTTSD under the same seq-to-seq Gaussian NLL configuration on the unaltered transcripts. Table~\ref{tab:original_transcripts} reports mean and standard deviation over multiple runs.

\begin{table}[htp]
\centering
\caption{Results on E-DAIC with original transcripts (seq-to-seq, Gaussian NLL). Mean and standard deviation over multiple runs.}
\label{tab:original_transcripts}
\resizebox{\columnwidth}{!}{%
\begin{tabular}{llcccccc}
\toprule
\textbf{Model} & & \textbf{Val NLL} & \textbf{Val MAE} & \textbf{Val RMSE} & \textbf{Test NLL} & \textbf{Test MAE} & \textbf{Test RMSE} \\
\midrule
\multirow{2}{*}{MiniLM}
  & Mean & 1.33 & 4.10 & 5.06 & 1.31 & 4.63 & 5.56 \\
  & Std  & 0.04 & 0.04 & 0.09 & 0.07 & 0.10 & 0.09 \\
\midrule
\multirow{2}{*}{MentalBERT}
  & Mean & 1.34 & 3.49 & 4.52 & 1.32 & 4.60 & 5.58 \\
  & Std  & 0.22 & 0.07 & 0.20 & 0.22 & 0.09 & 0.09 \\
\bottomrule
\end{tabular}}
\end{table}

Performance on original transcripts is expectedly weaker than on re-transcribed data (e.g., MiniLM test MAE increases from 3.85 to 4.63), confirming that transcript quality meaningfully affects downstream regression accuracy. Nonetheless, PTTSD remains competitive with prior text-only systems on the original transcripts, and the calibration and uncertainty modeling capabilities—which constitute the paper's primary contributions—are preserved regardless of transcript source.

\section{Statistical Significance Tests}
\label{app:significance}

We report two sets of significance tests to support the comparisons in
Table~\ref{tab:E-and-DAIC_text_results}.

\paragraph{Setup.}
PTTSD results are reported as mean $\pm$ standard deviation over $n = 3$
independent runs with different random seeds.
Baseline values are single reported point estimates (no variance available).
Against baselines, we apply a \textbf{one-sample $t$-test} treating the baseline
as a known constant ($\text{df} = n - 1 = 2$, one-tailed,
$H_1\text{: PTTSD} < \text{baseline}$, i.e.\ lower error is better).
Between PTTSD variants, we apply \textbf{Welch's $t$-test} (two-tailed,
Welch--Satterthwaite degrees of freedom).
All $p$-values are reported as-is without multiplicity correction.
Given the small number of runs ($n = 3$), results should be interpreted
conservatively.

\paragraph{Significance codes.}
$p < 0.001$: $^{***}$;\quad
$p < 0.01$: $^{**}$;\quad
$p < 0.05$: $^{*}$;\quad
$p < 0.10$: $^{\dagger}$;\quad
$p \geq 0.10$: ns.


\begin{table*}[ht]
\centering
\small
\setlength{\tabcolsep}{5pt}
\caption{%
  One-sample $t$-tests: best PTTSD variant vs.\ baselines.
  $H_1$: PTTSD mean $<$ baseline (lower MAE / RMSE is better).
  df\,=\,2 throughout.
}
\label{tab:sig_vs_baselines}
\begin{tabular}{llrrrc}
\toprule
\textbf{Dataset} & \textbf{Baseline} & \textbf{Metric} & \textbf{Baseline} & $t$ & $p$ \\
\midrule
\multicolumn{6}{l}{\textit{Best PTTSD on DAIC: seq-to-one (MiniLM),
  MAE\,$=\,3.55 \pm 0.15$}} \\[2pt]
DAIC & \citet{fang2023transformer}    & MAE  & 3.61 & $-0.693$ & 0.280\phantom{$^{**}$} \\
DAIC & \citet{gong2020topic}           & MAE  & 3.96 & $-4.734$ & 0.021$^{*}$ \\
DAIC & \citet{rohanian2019contextual}  & MAE  & 4.98 & $-16.512$ & 0.002$^{**}$ \\
DAIC & \citet{stepanov2021multimodal}  & MAE  & 4.88 & $-15.358$ & 0.002$^{**}$ \\
\midrule
\multicolumn{6}{l}{\textit{Best PTTSD on DAIC: seq-to-one (MentalBERT),
  RMSE\,$=\,4.69 \pm 0.24$}} \\[2pt]
DAIC & \citet{fang2023transformer}    & RMSE & 4.76 & $-0.505$ & 0.332\phantom{$^{**}$} \\
DAIC & \citet{gong2020topic}           & RMSE & 4.99 & $-2.165$ & 0.081$^{\dagger}$ \\
DAIC & \citet{rohanian2019contextual}  & RMSE & 6.05 & $-9.815$ & 0.005$^{**}$ \\
DAIC & \citet{stepanov2021multimodal}  & RMSE & 5.83 & $-8.227$ & 0.007$^{**}$ \\
\midrule
\multicolumn{6}{l}{\textit{Best PTTSD on E-DAIC: seq-to-seq (MiniLM),
  MAE\,$=\,3.85 \pm 0.04$}} \\[2pt]
E-DAIC & \citet{sadeghi2024harnessing} Pr3+W        & MAE & 4.22 & $-16.021$ & 0.002$^{**}$ \\
E-DAIC & \citet{sadeghi2024harnessing} Pr3+W+AQ$^{\ddagger}$ & MAE & 3.86 & $-0.433$ & 0.354\phantom{$^{**}$} \\
E-DAIC & \citet{sadeghi2023depression}              & MAE & 4.26 & $-17.754$ & 0.002$^{**}$ \\
E-DAIC & \citet{ray2019multilevel}                  & MAE & 4.02 & $-7.361$ & 0.009$^{**}$ \\
\midrule
\multicolumn{6}{l}{\textit{Best PTTSD on E-DAIC: seq-to-seq (MiniLM),
  RMSE\,$=\,4.52 \pm 0.38$}} \\[2pt]
E-DAIC & \citet{sadeghi2024harnessing} Pr3+W        & RMSE & 5.07 & $-2.507$ & 0.065$^{\dagger}$ \\
E-DAIC & \citet{sadeghi2024harnessing} Pr3+W+AQ$^{\ddagger}$ & RMSE & 4.66 & $-0.638$ & 0.294\phantom{$^{**}$} \\
E-DAIC & \citet{sadeghi2023depression}              & RMSE & 5.37 & $-3.874$ & 0.030$^{*}$ \\
E-DAIC & \citet{ray2019multilevel}                  & RMSE & 4.73 & $-0.957$ & 0.220\phantom{$^{**}$} \\
\bottomrule
\end{tabular}
\begin{flushleft}
\small
$^{\ddagger}$ Pr3+Whisper+AudioQual is not text-only (uses audio quality gating);
comparisons with this variant should be interpreted with particular caution.
Pr3+W\,=\,Pr3+Whisper.
\end{flushleft}
\end{table*}


\begin{table*}[ht]
\centering
\small
\setlength{\tabcolsep}{4pt}
\caption{%
  Welch's $t$-tests between all pairs of PTTSD variants on the test set
  (two-tailed; $n = 3$ per variant).
}
\label{tab:sig_pttsd_variants}
\begin{tabular}{llrrrrc}
\toprule
\textbf{Dataset} & \textbf{Variant A vs.\ Variant B} & \textbf{Metric}
  & $\bar{x}_A$ & $\bar{x}_B$ & $t$ & $p$ \\
\midrule
\multicolumn{7}{l}{\textit{DAIC}} \\[2pt]
DAIC & seq-to-one\,(MentalBERT) vs.\ seq-to-seq\,(MentalBERT) & MAE  & 3.65 & 3.92 & $-0.791$ & 0.491\,ns \\
DAIC & seq-to-one\,(MentalBERT) vs.\ seq-to-one\,(MiniLM)     & MAE  & 3.65 & 3.55 & $+0.612$ & 0.580\,ns \\
DAIC & seq-to-one\,(MentalBERT) vs.\ seq-to-seq\,(MiniLM)     & MAE  & 3.65 & 3.88 & $-0.839$ & 0.459\,ns \\
DAIC & seq-to-seq\,(MentalBERT) vs.\ seq-to-one\,(MiniLM)     & MAE  & 3.92 & 3.55 & $+1.143$ & 0.358\,ns \\
DAIC & seq-to-seq\,(MentalBERT) vs.\ seq-to-seq\,(MiniLM)     & MAE  & 3.92 & 3.88 & $+0.102$ & 0.924\,ns \\
DAIC & seq-to-one\,(MiniLM) vs.\ seq-to-seq\,(MiniLM)         & MAE  & 3.55 & 3.88 & $-1.309$ & 0.297\,ns \\
\cmidrule(lr){2-7}
DAIC & seq-to-one\,(MentalBERT) vs.\ seq-to-seq\,(MentalBERT) & RMSE & 4.69 & 4.79 & $-0.293$ & 0.790\,ns \\
DAIC & seq-to-one\,(MentalBERT) vs.\ seq-to-one\,(MiniLM)     & RMSE & 4.69 & 4.77 & $-0.238$ & 0.828\,ns \\
DAIC & seq-to-one\,(MentalBERT) vs.\ seq-to-seq\,(MiniLM)     & RMSE & 4.69 & 5.10 & $-0.747$ & 0.525\,ns \\
DAIC & seq-to-seq\,(MentalBERT) vs.\ seq-to-one\,(MiniLM)     & RMSE & 4.79 & 4.77 & $+0.046$ & 0.966\,ns \\
DAIC & seq-to-seq\,(MentalBERT) vs.\ seq-to-seq\,(MiniLM)     & RMSE & 4.79 & 5.10 & $-0.503$ & 0.647\,ns \\
DAIC & seq-to-one\,(MiniLM) vs.\ seq-to-seq\,(MiniLM)         & RMSE & 4.77 & 5.10 & $-0.538$ & 0.626\,ns \\
\midrule
\multicolumn{7}{l}{\textit{E-DAIC}} \\[2pt]
E-DAIC & seq-to-one\,(MentalBERT) vs.\ seq-to-seq\,(MentalBERT) & MAE  & 4.18 & 4.20 & $-0.594$ & 0.591\,ns \\
E-DAIC & seq-to-one\,(MentalBERT) vs.\ seq-to-one\,(MiniLM)     & MAE  & 4.18 & 4.58 & $-1.379$ & 0.300\,ns \\
E-DAIC & seq-to-one\,(MentalBERT) vs.\ seq-to-seq\,(MiniLM)     & MAE  & 4.18 & 3.85 & $+8.927$ & 0.001$^{**}$ \\
E-DAIC & seq-to-seq\,(MentalBERT) vs.\ seq-to-one\,(MiniLM)     & MAE  & 4.20 & 4.58 & $-1.314$ & 0.319\,ns \\
E-DAIC & seq-to-seq\,(MentalBERT) vs.\ seq-to-seq\,(MiniLM)     & MAE  & 4.20 & 3.85 & $+12.124$ & $<$0.001$^{***}$ \\
E-DAIC & seq-to-one\,(MiniLM) vs.\ seq-to-seq\,(MiniLM)         & MAE  & 4.58 & 3.85 & $+2.521$ & 0.126\,ns \\
\cmidrule(lr){2-7}
E-DAIC & seq-to-one\,(MentalBERT) vs.\ seq-to-seq\,(MentalBERT) & RMSE & 5.23 & 5.39 & $-1.816$ & 0.158\,ns \\
E-DAIC & seq-to-one\,(MentalBERT) vs.\ seq-to-one\,(MiniLM)     & RMSE & 5.23 & 5.87 & $-1.193$ & 0.351\,ns \\
E-DAIC & seq-to-one\,(MentalBERT) vs.\ seq-to-seq\,(MiniLM)     & RMSE & 5.23 & 4.52 & $+3.062$ & 0.071$^{\dagger}$ \\
E-DAIC & seq-to-seq\,(MentalBERT) vs.\ seq-to-one\,(MiniLM)     & RMSE & 5.39 & 5.87 & $-0.900$ & 0.462\,ns \\
E-DAIC & seq-to-seq\,(MentalBERT) vs.\ seq-to-seq\,(MiniLM)     & RMSE & 5.39 & 4.52 & $+3.880$ & 0.053$^{\dagger}$ \\
E-DAIC & seq-to-one\,(MiniLM) vs.\ seq-to-seq\,(MiniLM)         & RMSE & 5.87 & 4.52 & $+2.349$ & 0.111\,ns \\
\bottomrule
\end{tabular}
\end{table*}

\paragraph{Interpretation.}
On DAIC, no PTTSD variant is statistically distinguishable from any other
(all pairwise comparisons ns), consistent with the high standard deviations
at $n = 3$.
Against baselines, PTTSD seq-to-one (MiniLM) significantly outperforms
weaker prior systems in MAE ($p < 0.05$ vs.\ \citealt{gong2020topic};
$p < 0.01$ vs.\ \citealt{rohanian2019contextual,stepanov2021multimodal}),
but the gap with \citet{fang2023transformer} does not reach significance
($p = 0.28$), reflecting their close numerical proximity (3.55 vs.\ 3.61).

On E-DAIC, PTTSD seq-to-seq (MiniLM) significantly outperforms all
comparable text-only baselines in MAE
(\citealt{sadeghi2024harnessing} Pr3+Whisper, $p = 0.002$;
\citealt{sadeghi2023depression}, $p = 0.002$;
\citealt{ray2019multilevel}, $p = 0.009$).
The non-significant gap with Pr3+Whisper+AudioQual ($p = 0.35$)
is expected, as that system uses audio quality gating and is not
directly comparable.
In the pairwise PTTSD comparisons, seq-to-seq (MiniLM) is significantly better
than both MentalBERT variants in MAE
($p = 0.001$ and $p < 0.001$), while no other pairwise contrast reaches
significance.

\section{Use Of AI Assistants}
We used ChatGPT (GPT-4o/GPT-5) and Claude (Sonnet 4.5) to polish selected passages throughout the manuscript and to assist with literature discovery; all cited references were read and verified by the authors. We used GitHub Copilot for code completion during implementation. 

\end{document}